\documentclass[a4paper]{article}

\usepackage{amssymb}
\usepackage{graphicx}
\usepackage{multirow}
\usepackage{afterpage}
\usepackage{booktabs}

\usepackage{array}
\newcolumntype{P}[1]{>{\centering\arraybackslash}p{#1}}
\newcolumntype{M}[1]{>{\centering\arraybackslash}m{#1}}

\def\email#1{\texttt{#1}}

\def\institute#1{#1}

\usepackage{url}
\urldef{\mailsa}\path|yshean@1utar.my|
\urldef{\mailsb}\path|tayyh@utar.edu.my|

\begin{document}

%%\mainmatter  % start of an individual contribution

\title{Abnormal Event Detection in Videos\\ using Spatiotemporal Autoencoder}

%%\titlerunning{Abnormal Event Detection in Videos using Spatiotemporal Autoencoder}

\author{Yong Shean Chong \qquad Yong Haur Tay \\ \email{yshean@1utar.my} \qquad \email{tayyh@utar.edu.my} \\\\
\institute{Lee Kong Chian Faculty of Engineering Science,\\
	Universiti Tunku Abdul Rahman, 43000 Kajang, Malaysia.}
}
%
%%\authorrunning{Yong Shean Chong and Yong Haur Tay}


%%\toctitle{Lecture Notes in Computer Science}
%%\tocauthor{Authors' Instructions}
\maketitle

\begin{abstract}
We present an efficient method for detecting anomalies in videos. Recent applications of convolutional neural networks have shown promises of convolutional layers for object detection and recognition, especially in images. However, convolutional neural networks are supervised and require labels as learning signals. We propose a spatiotemporal architecture for anomaly detection in videos including crowded scenes. Our architecture includes two main components, one for spatial feature representation, and one for learning the temporal evolution of the spatial features. Experimental results on Avenue, Subway and UCSD benchmarks confirm that the detection  accuracy of our method is comparable to state-of-the-art methods at a considerable speed of up to 140 fps.
%%\keywords{Video anomaly detection, spatiotemporal feature %%learning, regularity, autoencoder}
\end{abstract}

\section{Introduction}

With the rapid growth of video data, there is an increasing need not only for recognition of objects and their behaviour, but in particular for detecting the rare, interesting occurrences of unusual objects or suspicious behaviour in the large body of ordinary data. Finding such abnormalities in videos is crucial for applications ranging from automatic quality control to visual surveillance.

Meaningful events that are of interest in long video sequences, such as surveillance footage, often have an extremely low probability of occurring. As such, manually detecting such events, or anomalies, is a very meticulous job that often requires more manpower than is generally available. This has prompted the need for automated detection and segmentation of sequences of interest. However, present technology requires an enormous amount of configuration efforts on each video stream prior to the deployment of the video analysis process, even with that, those events are based on some predefined heuristics, which makes the detection model difficult to generalize to different surveillance scenes. 

Video data is challenging to represent and model due to its high dimensionality, noise, and a huge variety of events and interactions. Anomalies are also highly contextual, for example, running in a restaurant would be an anomaly, but running at a park would be normal. Moreover, the definition of anomaly can be ambiguous and often vaguely defined. A person may think walking around on a subway platform is normal, but some may think it should be flagged as an anomaly since it could be suspicious. These challenges have made it difficult for machine learning methods to identify video patterns that produce anomalies in real-world applications.

There are many successful cases in the related field of action recognition \cite{tran2016,jia2014,ji2013,oneata2013}. However, these methods only applicable to labelled video footages where events of interest are clearly defined and does not involve highly occluded scenes, such as crowded scenes. Furthermore, the cost of labelling every type of event is extremely high. Even so, it is not guaranteed to cover every past and future events. The recorded video footage is likely not long enough to capture all types of activities, especially abnormal activities which rarely or never occurred.

Recent effort on detecting anomalies by treating the task as a binary classification problem (normal and abnormal) \cite{zhou2016} proved it being effective and accurate, but the practicality of such method is limited since footages of abnormal events are difficult to obtain due to its rarity. Therefore, many researchers have turned to models that can be trained using little to no supervision, including spatiotemporal features \cite{lu2013,zhao2011}, dictionary learning \cite{yen2013} and autoencoders \cite{sabokrou2015}. Unlike supervised methods, these methods only require unlabelled video footages which contain little or no abnormal event, which are easy to obtain in real-world applications. A description of these methodologies and their limitations are discussed in the next section. 

This paper presents a novel framework to represent video data by a set of general features, which are inferred automatically from a long video footage through a deep learning approach. Specifically, a deep neural network composed of a stack of convolutional autoencoders was used to process video frames in an unsupervised manner that captured spatial structures in the data, which, grouped together, compose the video representation. Then, this representation is fed into a stack of convolutional temporal autoencoders to learn the regular temporal patterns.

Our proposed method is domain free (i.e., not related to any specific task, no domain expert required), does not require any additional human effort, and can be easily applied to different scenes. To prove the effectiveness of the proposed method we apply the method to real-world datasets and show that our method consistently outperforms similar methods while maintaining a short running time.

\subsection{Our Contributions}
The main characteristics of our approach and also the contributions of this research are as follows:
\begin{itemize}
	\item We wish to reduce the labor-intensive effort in feature engineering that results in a representation of the data that can support effective machine learning. This can be done by replacing low-level handcrafted features with learned hierarchical features. With the help of autoencoders, we are able to find representative features by learning from data instead of forming suitable features based on our knowledge.\\
	\item We replace traditional sparse coding methods with autoencoders. Unlike existing methods, there is no separation between extracting feature representation of videos and learning a model of features. In addition, by having multiple layers of hidden units in autoencoder, hierarchical feature learning can be achieved.
\end{itemize} 

\section{Related Work}

Most of these abnormal instances are beforehand unknown, as this would require predicting all the ways something could happen out of the norm. It is therefore simply impossible to learn a model for all that is abnormal or irregular. But how can we find an anomaly without what to look for? 

Since it is easier to get video data where the scene is normal in contrast to obtaining what is abnormal, we could focus on a setting where the training data contains only normal visual patterns. A popular approach adopted by researchers in this area is to first learn the normal patterns from the training videos, then anomalies are detected as events deviated from the normal patterns \cite{lu2013,cong2011,zhao2011,li2011}. The majority of the work on anomaly detection relies on the extraction of local features from videos, that are then used to train a normalcy model. 

Trajectories have long been popular in video analysis and anomaly detection \cite{zhou2015,li2011,piciarelli2008,mo2014}.  A common characteristic of trajectory-based approaches is the deviation of nominal classes of object trajectories in a training phase, and the comparison of new test trajectories against the nominal classes in an evaluation phase. A statistically significant deviation from all classes indicates an anomaly. However, the accuracy of trajectory analysis relies heavily on tracking, which precise tracking still remains a significant challenge in computer vision, particularly in complex situations. Tracking-based approaches are suitable for scenes with few objects but are impractical for detecting abnormal patterns in a crowded or complex scene.

Non-tracking approaches that focus on spatiotemporal anomalies in videos also exist. These rely mainly on extracting and analyzing local low-level visual features, such as the histogram of oriented gradients \cite{xiao2015}, the histogram of oriented flows \cite{laptev2008} and optical flow \cite{reddy2011}, by employing spatiotemporal video volumes (dense sampling or interest point selection) \cite{dollar2005}. These local features are then grouped in clusters, i.e., bags of visual words (BOV), according to similarity metrics. Their popularity is due to their low computational cost, as well as their ability to focus on abnormal behaviour, even in extremely crowded scenes \cite{kratz2009}. Another similar technique is sparse reconstruction \cite{cong2011,zhao2011}. The fundamental underlying assumption of these methods is that any new feature representation of a normal/anomalous event can be approximately modeled as a (sparse) linear combination of feature representations (of previously observed events) in a trained dictionary. This assumes that all previously observed events are normal events. 

However, since classical BOV approaches group similar volumes (summarize), they destroy all compositional information in the process of grouping visual words. It is also required to pre-determine the number of clusters, which can only be found through trial-and-error during testing time. In addition, codebook models require searching over a large space \cite{roshtkhari2013} even during the time of testing, making it impractical for real-time anomaly detection. 

The success of deep learning methods in various applications consequently caused the rise of such methods in anomaly detection. The term deep learning refers to learning a hierarchical set of features through multiple layers of hidden nodes in an artificial neural network. Unlike previously stated methods, there is no need to define a specific set of features to extract from the dataset -- deep learning methods learn the useful features directly from the data with minimal preprocessing. Specifically, convolutional neural networks (ConvNet) have proved its effectiveness in a wide range of applications such as object recognition \cite{simonyan2014obj}, person detection \cite{vu2016}, and action recognition \cite{tran2016,simonyan2014vid}. ConvNet consists of a stack of convolutional layers with a fully-connected layer and a softmax classifier, and convolutional autoencoder is essentially a ConvNet with its fully-connected layer and classifier replaced by a mirrored stack of convolutional layers. The authors of \cite{zhou2016} applied a 3D ConvNet on classifying anomalies, whereas \cite{hasan2016} used an end-to-end convolutional autoencoder to detect anomalies in surveillance videos. Their reported result proves the usefulness of learned representation on videos through a stack of convolutional layers. On the other hand, long short term memory (LSTM) model is well-known for learning temporal patterns and predicting time series data. \cite{medel2016} has recently proposed to apply convolutional LSTMs for learning the regular temporal patterns in videos and his findings show great promise of what deep neural network can learn.

Despite its simplicity, some limitations remain in these recently proposed methods. Though 3D ConvNet performed excellently in learning discriminative features between the anomalies and the normal events, it is impractical to apply in real-world scenarios due to the absence of video segments containing abnormal events. Meanwhile, in the convolutional autoencoder proposed by \cite{hasan2016}, convolution and pooling operations are performed only spatially, even though the proposed network takes multiple frames as input, because of the 2D convolutions, after the first convolution layer, temporal information is collapsed completely \cite{tran2016}. Besides, convolutional LSTM layers applied by \cite{medel2016} are memory-intensive -- the training will need to be executed on very small mini-batches, which results in slow training and testing time.

\section{Methodology}

The method described here is based on the principle that when an abnormal event occurs, the most recent frames of video will be significantly different than the older frames. Inspired by \cite{hasan2016}, we train an end-to-end model that consists of a spatial feature extractor and a temporal encoder-decoder which together learns the temporal patterns of the input volume of frames. The model is trained with video volumes consists of only normal scenes, with the objective to minimize the reconstruction error between the input video volume and the output video volume reconstructed by the learned model. After the model is properly trained, normal video volume is expected to have low reconstruction error, whereas video volume consisting of abnormal scenes is expected to have high reconstruction error. By thresholding on the error produced by each testing input volumes, our system will be able to detect when an abnormal event occurs.

Our approach consists of three main stages:

\subsection{Preprocessing}

The task of this stage is to convert raw data to the aligned and acceptable input for the model. Each frame is extracted from the raw videos and resized to $227\times227$. To ensure that the input images are all on the same scale, the pixel values are scaled between 0 and 1 and subtracted every frame from its global mean image for normalization. The mean image is calculated by averaging the pixel values at each location of every frame in the training dataset. After that, the images are converted to grayscale to reduce dimensionality. The processed images are then normalized to have zero mean and unit variance. 

The input to the model is video volumes, where each volume consists of 10 consecutive frames with various skipping strides. As the number of parameters in this model is large, large amount of training data is needed. Following \cite{hasan2016}’s practice, we perform data augmentation in the temporal dimension to increase the size of the training dataset. To generate these volumes, we concatenate frames with stride-1, stride-2, and stride-3. For example, the first stride-1 sequence is made up of frame \{1, 2, 3, 4, 5, 6, 7, 8, 9, 10\}, whereas the first stride-2 sequence contains frame number \{1, 3, 5, 7, 9, 11, 13, 15, 17, 19\}, and stride-3 sequence would contain frame number \{1, 4, 7, 10, 13, 16, 19, 22, 25, 28\}. Now the input is ready for model training.

\subsection{Feature Learning}

\begin{figure}
	\centering
	\includegraphics[width=\textwidth]{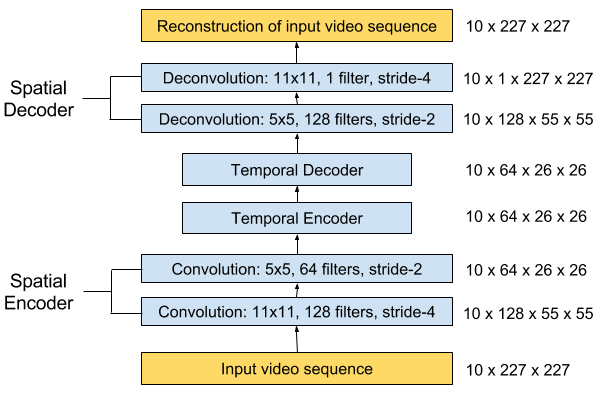}
	\caption{Our proposed network architecture. It takes a sequence of length T as input, and output a reconstruction of the input sequence. The numbers at the rightmost denote the output size of each layer. The spatial encoder takes one frame at a time as input, after which $T=10$ frames have been processed, the encoded features of 10 frames are concatenated and fed into temporal encoder for motion encoding. The decoders mirror the encoders to reconstruct the video volume.}
	\label{fig:ours_spatial}
\end{figure}

\begin{figure}
	\centering
	\includegraphics[width=4.5cm]{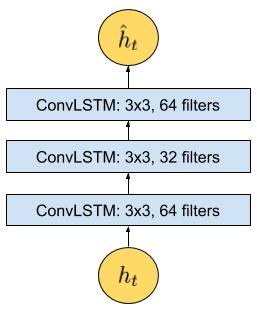}
	\caption{The zoomed-in architecture at time $t$, where $t$ is the input vector at this time step. The temporal encoder-decoder model has 3 convolutional LSTM (ConvLSTM) layers. }
	\label{fig:ours_temporal}
\end{figure}

We propose a convolutional spatiotemporal autoencoder to learn the regular patterns in the training videos. Our proposed architecture consists of two parts --- spatial autoencoder for learning spatial structures of each video frame, and temporal encoder-decoder for learning temporal patterns of the encoded spatial structures. As illustrated in Figure \ref{fig:ours_spatial} and \ref{fig:ours_temporal}, the spatial encoder and decoder have two convolutional and deconvolutional layers respectively, while the temporal encoder is a three-layer convolutional long short term memory (LSTM) model. Convolutional layers are well-known for its superb performance in object recognition, while LSTM model is widely used for sequence learning and time-series modelling and has proved its performance in applications such as speech translation and handwriting recognition.

\subsubsection{Autoencoder}

Autoencoders, as the name suggests, consist of two stages: encoding and decoding. It was first used to reduce dimensionality by setting the number of encoder output units less than the input. The model is usually trained using back-propagation in an unsupervised manner,
by minimizing the reconstruction error of the decoding results from the original inputs. With the activation function chosen to be nonlinear, an autoencoder can extract more useful features than some common linear transformation methods such as PCA.

\subsubsection{Spatial Convolution}

The primary purpose of convolution in case of a convolutional network is to extract features from the input image. Convolution preserves the spatial relationship between pixels by learning image features using small squares of input data. Mathematically, convolution operation performs dot products between the filters and local regions of the input. Suppose that we have some $n \times n$ square input layer which is followed by the convolutional layer. If we use an $m \times m$ filter $W$, the convolutional layer output will be of size $(n-m+1) \times (n-m+1)$. 

A convolutional network learns the values of these filters on its own during the training process, although we still need to specify parameters such as the number of filters, filter size, the number of layers before training. With more number of filters we have, more image features get extracted and the better the network becomes at recognizing patterns in unseen images. However, more filters would add to computational time and exhaust memory faster, so we need to find balance by not setting the number of filters too large.

\subsubsection{Recurrent Neural Network (RNN)}

In a traditional feedforward neural network, we assume that all inputs (and outputs) are independent of each other. However, learning temporal dependencies between inputs are important in tasks involving sequences, for example, a word predictor model should be able to derive information from the past inputs. RNN works just like a feedforward network, except that the values of its output vector are influenced not only by the input vector but also on the entire history of inputs. In theory, RNNs can make use of information in arbitrarily long sequences, but in practice, they are limited to looking back only a few steps due to vanishing gradients. 

\subsubsection{Long Short Term Memory (LSTM)}

\begin{figure}
	\centering
	\includegraphics[height=6.2cm]{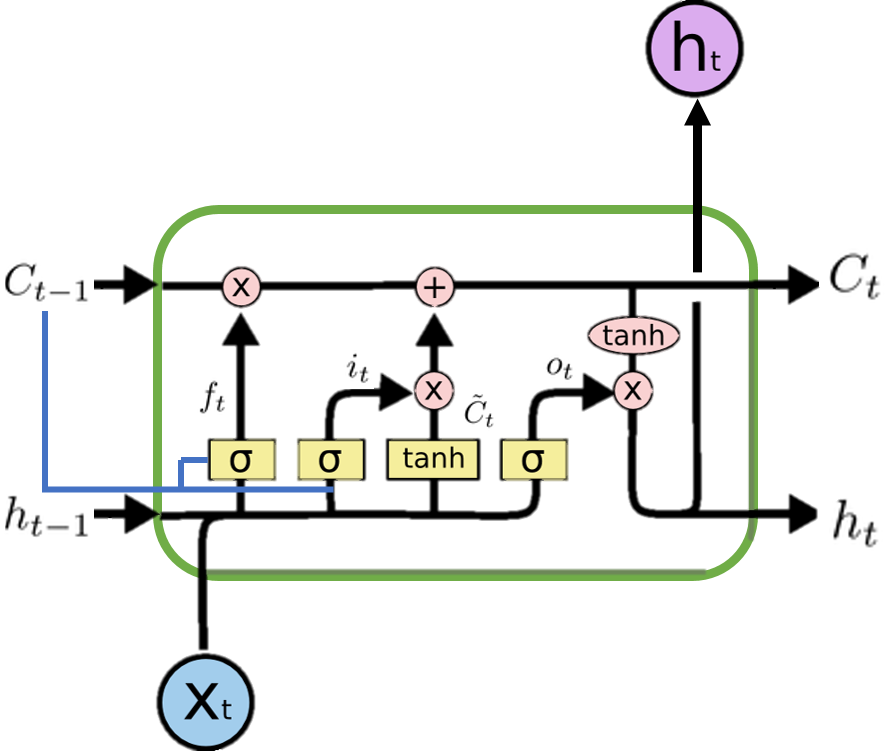}
	\caption{The structure of a typical LSTM unit. The blue line represents an optional ‘peephole’ structure, which allows the internal state to look back (‘peep’) at the previous cell state $C_{t-1}$ for a better decision. Best viewed in colour.}
	\label{fig:lstm}
\end{figure}

To overcome this problem, a variant of RNN is introduced: long short term memory (LSTM) model which incorporates a recurrent gate called forget gate. With the new structure, LSTMs prevent backpropagated errors from vanishing or exploding, thus can work on long sequences and they can be stacked together to capture higher level information. The formulation of a typical LSTM unit is summarized with Figure \ref{fig:lstm} and equations (1) through (6).

\begin{equation}
f_t = \sigma(W_f \otimes [h_{t-1}, x_t] + b_f)
\end{equation}

\begin{equation}
i_t = \sigma(W_i \otimes [h_{t-1}, x_t] + b_i)
\end{equation}

\begin{equation}
\hat{C_t} = tanh(W_C \otimes [h_{t-1}, x_t] + b_C)
\end{equation}

\begin{equation}
C_t = f_t \otimes C_{t-1} + i_t \otimes \hat{C_t}
\end{equation}

\begin{equation}
o_t = \sigma(W_o \otimes [h_{t-1}, x_t] + b_o)
\end{equation}

\begin{equation}
h_t = o_t \otimes tanh(C_t)
\end{equation}

Equation (1) represents the forget layer, (2) and (3) are where new information is added, (4) combines old and new information, whereas (5) and (6) output what has been learned so far to the LSTM unit at the next timestep. The variable $x_t$ denotes the input vector, $h_t$ denotes the hidden state, and $C_t$ denotes the cell state  at time $t$. $W$ are the trainable weight matrices, $b$ are the bias vectors, and the symbol $\otimes$ denotes the Hadamard product.  

\subsubsection{Convolutional LSTM}
A variant of the LSTM architecture, namely Convolutional Long Short-term Memory (ConvLSTM) model was introduced by Shi et al. in \cite{shi2015} and has been recently utilized by Patraucean et al. in \cite{patraucean2016} for video frame prediction. Compared to the usual fully connected LSTM (FC-LSTM), ConvLSTM has its matrix operations replaced with convolutions. By using convolution for both input-to-hidden and hidden-to-hidden connections, ConvLSTM requires fewer weights and yield better spatial feature maps. The formulation of the ConvLSTM unit can be summarized with (7) through (12). 

\begin{equation}
f_t = \sigma(W_f \ast [h_{t-1}, x_t, C_{t-1}] + b_f)
\end{equation}

\begin{equation}
i_t = \sigma(W_i \ast [h_{t-1}, x_t, C_{t-1}] + b_i)
\end{equation}

\begin{equation}
\hat{C_t} = tanh(W_C \ast [h_{t-1}, x_t] + b_C)
\end{equation}

\begin{equation}
C_t = f_t \otimes C_{t-1} + i_t \otimes \hat{C_t}
\end{equation}

\begin{equation}
o_t = \sigma(W_o \ast [h_{t-1}, x_t, C_{t-1}] + b_o)
\end{equation}

\begin{equation}
h_t = o_t \otimes tanh(C_t)
\end{equation}

While the equations are similar in nature to (1) through (6), the input is fed in as images, while the set of weights for every connection is replaced by convolutional filters (the symbol $\ast$ denotes a convolution operation). This allows ConvLSTM work better with images than the FC-LSTM due to its ability to propagate spatial characteristics temporally through each ConvLSTM state. Note that this convolutional variant also adds an optional 'peephole' connections to allow the unit to derive past information better.

\subsection{Regularity Score}
Once the model is trained, we can evaluate our model’s performance by feeding in testing data and check whether it is capable of detecting abnormal events while keeping false alarm rate low. To better compare with \cite{hasan2016}, we used the same formula to calculate the regularity score for all frames, the only difference being the learned model is of a different kind. The reconstruction error of all pixel values I in frame t of the video sequence is taken as the Euclidean distance between the input frame and the reconstructed frame:

\begin{equation}
e(t) = ||x(t) - f_W(x(t))||_2
\end{equation}

where $f_W$ is the learned weights by the spatiotemporal model. We then compute the abnormality score $s_a(t)$ by scaling  between 0 and 1. Subsequently, regularity score $s_r(t)$ can be simply derived by subtracting abnormality score from 1:

\begin{equation}
s_a(t) = \frac{e(t) - e(t)_{min}}{e(t)_{max}}
\end{equation}

\begin{equation}
s_r(t) = 1 - s_a(t)
\end{equation}

\subsection{Anomaly Detection}

\subsubsection{Thresholding} It is straightforward to determine whether a video frame is normal or anomalous. The reconstruction error of each frame determines whether the frame is classified as anomalous. The threshold determines how sensitive we wish the detection system to behave --- for example, setting a low threshold makes the system become sensitive to the happenings in the scene, where more alarms would be triggered. We obtain the true positive and false positive rate by setting at different error threshold in order to calculate the area under the receiver operating characteristic (ROC) curve (AUC). The equal error rate (EER) is obtained when false positive rate equals to the false negative rate.

\subsubsection{Event count} Following the practice in \cite{hasan2016}, to reduce the noisy and unmeaningful minima in the regularity score, we applied Persistence1D \cite{pers1d} algorithm to group local minima with a fixed temporal window of 50 frames. We assume local minima within 50 frames belong to the same abnormal event. This is a reasonable length of the temporal window as an abnormal event should be at least 2-3 seconds long to be meaningful (videos are captured at 24-25 fps).

\section{Experiments}

\subsection{Datasets}

We train our model on five most commonly used benchmarking datasets: Avenue \cite{lu2013}, UCSD Ped1 and Ped2 \cite{mahadevan2010}, Subway entrance and exit datasets \cite{adam2008}. All videos are taken from a fixed position for each dataset. All training videos contain only normal events. Testing videos have both normal and abnormal events. 

In Avenue dataset, there are total 16 training and 21 testing video clips. Each clip’s duration vary between less than a minute to two minutes long. The normal scenes consist of people walking between staircase and subway entrance, whereas the abnormal events are people running, walking in opposite direction, loitering and etc. The challenges of this dataset include camera shakes and a few outliers in the training data. Also, some normal pattern seldom appears in the training data.

UCSD Ped1 dataset has 34 training and 36 testing video clips, where each clip contains 200 frames. The videos consist of groups of people walking towards and away from the camera. UCSD Ped2 dataset has 16 training and 12 testing video clips, where the number of frames of each clip varies. The videos consist of walking pedestrians parallel to the camera plane. Anomalies of the two datasets include bikers, skaters, carts, wheelchairs and people walking in the grass area. 

Subway entrance dataset is 1 hour 36 minutes long with 66 unusual events of five different types: walking in the wrong direction (WD), no payment (NP), loitering (LT), irregular interactions between people (II), and miscellaneous (e.g. sudden stop, running fast). First 20 minutes of the video is used for training.

Subway exit dataset is 43 minutes long with 19 unusual events of three types: walking in the wrong direction (WD), loitering (LT), and miscellaneous (e.g. sudden stop, looking around, a janitor cleaning the wall, gets off the train and gets on the train again quickly. First 5 minutes of the video is used for training.

\subsection{Model Parameters}

We train the model by minimizing the reconstruction error of the input volume. We use Adam optimizer to allow it taking the role of setting the learning rate automatically based on the model’s weight update history. We use mini-batches of size 64 and each training volume is trained for a maximum of 50 epochs or until the reconstruction loss of validation data stop decreasing after 10 consecutive epochs. Hyperbolic tangent is chosen as the activation function of spatial encoder and decoder. To ensure the symmetry of the encoding and decoding function, we did not use rectified linear unit (ReLU) despite its regularization ability because activated values from ReLU have no upper bound. 

\subsection{Results and Analysis}

\subsubsection{Quantitative Analysis: ROC and Anomalous Event Count}

% \usepackage{multirow}
% \usepackage{graphicx}
\begin{table}[]
	\centering
	\caption{Comparison of area under ROC curve (AUC) and Equal Error Rate (EER) of different methods. Higher AUC and lower EER are better. Most papers did not publish their AUC/EER for avenue, subway entrance and exit dataset.}
	\label{table:auc_eer}
	%\resizebox{\textwidth}{!}{%
		%\fontsize{4}{5}\selectfont
		%\small
		\begin{tabular}{M{2.3cm}M{1.8cm}M{1.8cm}M{1.8cm}M{1.8cm}M{1.8cm}}
			\hline
			\multirow{2}{*}{Method} & \multicolumn{5}{c}{AUC/EER (\%)}                                                                                                                        \\ \cline{2-6} 
			& Ped1      & Ped2      & Avenue    & \begin{tabular}[c]{@{}c@{}}Subway \\ Entrance\end{tabular} & \begin{tabular}[c]{@{}c@{}}Subway \\ Exit\end{tabular} \\ \cline{1-6}
			Adam \cite{adam2008}                    & 77.1/38.0 & -/42.0    & \multicolumn{3}{c}{\multirow{4}{*}{N/A}}                                                                                        \\
			SF \cite{mehran2009}      & 67.5/31.0 & 55.6/42.0 & \multicolumn{3}{c}{}                                                                                                            \\
			MPPCA \cite{mahadevan2010}                   & 66.8/40.0 & 69.3/30.0 & \multicolumn{3}{c}{}                                                                                                            \\
			MPPCA+SF \cite{mahadevan2010}             & 74.2/32.0 & 61.3/36.0 & \multicolumn{3}{c}{}                                                                                                            \\
			HOFME \cite{wang2013}                     & 72.7/33.1    & 87.5/20.0    & N/A & 81.6/\textbf{22.8} & 84.9/17.8 \\ 
			ConvAE \cite{hasan2016}                  & 81.0/27.9 & \textbf{90.0}/21.7 & 70.2/25.1 & \textbf{94.3}/26.0                                                  & 80.7/9.9                                               \\
			Ours                    & \textbf{89.9}/\textbf{12.5} & 87.4/\textbf{12.0} & \textbf{80.3}/\textbf{20.7} & 84.7/23.7                                                  & \textbf{94.0}/\textbf{9.5}                                               \\ \hline
		\end{tabular}%
	%}
\end{table}

Table \ref{table:auc_eer} shows the frame-level AUC and EER of our and of other methods on all five datasets. We outperform all other considered methods in respect to frame-level EER. We also provide the event count comparison for Avenue dataset and the entrance and exit scenes in the Subway dataset in Table 2. For the entrance scenes, we are better than \cite{hasan2016} since we detect the same number of anomalies with less false alarms. For the exit scenes, we detected more abnormal events compared to \cite{hasan2016} but at the expense of higher false alarm rate.

% \usepackage{booktabs}
% \usepackage{multirow}
\begin{table}[]
	\centering
	\caption{Anomalous event and false alarm count detected by different methods. GT denotes groundtruth values of event count.}
	\label{table:event_count}
	%\resizebox{\textwidth}{!}{%
	%	\small
	\begin{tabular}{@{}M{3.2cm}M{3cm}M{2.5cm}M{2.5cm}@{}}
		\toprule
		\multirow{2}{*}{Method}     & \multicolumn{3}{c}{Anomalous Event Detected / False Alarm}                                                                                                                                                                      \\ \cline{2-4} 
		& \begin{tabular}[c]{@{}c@{}}Avenue \\ (GT: 47, \\ smaller set \\ GT: 14)\end{tabular} & \begin{tabular}[c]{@{}c@{}}Subway \\ Entrance \\ (GT: 66)\end{tabular} & \begin{tabular}[c]{@{}c@{}}Subway Exit \\ (GT: 19)\end{tabular} \\ \midrule
		Sparse combination \cite{lu2013} & 12/1 (smaller set)                                                                   & 57/4                                                                   & 19/2                                                            \\
		Space-time MRF \cite{kim2009}              & N/A                                                                                  & 56/3                                                                   & 18/0                                                            \\
		Online \cite{dutta2015}                         & N/A                                                                                  & 60/5                                                                   & 19/2                                                            \\
		ConvAE \cite{hasan2016}                      & 45/4                                                                                 & 61/15                                                                  & 17/5                                                            \\ %\cline{1-4}
		Ours                        & 44/6                                                                                 & 61/9                                                                   & 18/10                                                           \\ \bottomrule
	\end{tabular}
%}
\end{table}

The event count breakdown according to type of event is presented in Table \ref{table:avenue_count}, \ref{table:enter_count} and \ref{table:exit_count} for Avenue dataset, Subway entrance and exit datasets respectively. All throwing, loitering (LT) and irregular interaction (II) events are well captured by our proposed system. These are strong abnormalities that are significantly different from what was captured in the normal scenes. However, our system does have difficulties in detecting certain types of event. Missed detection of running and walking in opposite direction events are due to (1) the crowded activities where multiple foreground events take place; and (2) the object of interest is far away from the camera. Meanwhile, in Subway entrance and exit scenes, some wrong direction events are missed. On the other hand, some no payment (NP) events in Subway entrance scene are difficult to detect due to their similar motion compared to others walking through the barrier.

% \usepackage{booktabs}
% \usepackage{graphicx}
\begin{table}[]
	\centering
	\caption{Anomalous event and false alarm count detected by different methods on various event type in Avenue dataset.}
	\label{table:avenue_count}
	%\resizebox{\textwidth}{!}{%
		\begin{tabular}{@{}M{2cm}M{1.3cm}M{1.3cm}M{1.3cm}M{1.5cm}M{1.5cm}@{}}
			\toprule
			& Run & Loiter & Throw & Opposite Direction & False Alarm \\
			\midrule
			Groundtruth & 12 & 8 & 19 & 8 & 0 \\
			Ours & 10 & 8 & 19 & 7 & 12 \\
			\bottomrule
		\end{tabular}%
	%}
\end{table}

\begin{table}[]
	\centering
	\caption{Anomalous event and false alarm count detected by different methods on various event type in Subway Entrance dataset. WD: wrong direction; NP: no payment; LT: loitering; II: irregular interaction; Misc.: miscellaneous.}
	\label{table:enter_count}
	%\resizebox{\textwidth}{!}{%
		\begin{tabular}{@{}M{2cm}M{1cm}M{1cm}M{1cm}M{1cm}M{1cm}M{1.5cm}@{}}
			\toprule
			& WD & NP & LT & II & Misc. & False Alarm \\
			\midrule
			Groundtruth & 26 & 13 & 14 & 4 & 9 & 0 \\
			Ours & 24 & 10 & 14 & 4 & 9 & 9 \\
			\bottomrule
		\end{tabular}%
	%}
\end{table}

\begin{table}[]
	\centering
	\caption{Anomalous event and false alarm count detected by different methods on various event type in Subway Exit dataset. WD: wrong direction; LT: loitering; Misc.: miscellaneous.}
	\label{table:exit_count}
	%\resizebox{\textwidth}{!}{%
		\begin{tabular}{@{}M{2cm}M{1.5cm}M{1.5cm}M{1.5cm}M{1.5cm}@{}}
			\toprule
			& WD & LT & Misc. & False Alarm \\
			\midrule
			Groundtruth & 9 & 3 & 7 & 0 \\
			Ours & 8 & 3 & 7 & 10 \\
			\bottomrule
		\end{tabular}%
	%}
\end{table}

We also present a run-time analysis on our proposed abnormal event detection system, on CPU (Intel Xeon E5-2620) and GPU (NVIDIA Maxwell Titan X) respectively, in Table \ref{table:time}. The total time taken is well less than a quarter second per frame for both CPU and GPU configuration. Due to computational intensive multiplication operations when feeding the input through the convolutional autoencoders, it is recommended to run on GPU for a better speed of nearly 30 times faster than CPU. 

\begin{table}[]
	\centering
	\caption{Details of run-time during testing (second/frame).}
	\label{table:time}
	%\resizebox{\textwidth}{!}{%
		%\fontsize{4}{5}\selectfont
		%\small
		\begin{tabular}{@{}M{1.5cm}M{2.2cm}M{2.2cm}M{2.2cm}M{2.5cm}@{}}
			\toprule
			 & \multicolumn{4}{c}{Time (in sec)}\\
			 & Preprocessing & Representation & Classifying & Total \\
			\midrule
			CPU & 0.0010 & 0.2015 & 0.0002 & 0.2027 ($\sim$5fps) \\
			GPU & 0.0010 & 0.0058 & 0.0002 & 0.0070 ($\sim$143fps) \\
			\bottomrule
		\end{tabular}%
	%}
\end{table}

\subsubsection{Qualitative Analysis: Visualising Frame Regularity}

\begin{figure}
	\centering
	\includegraphics[width=\textwidth]{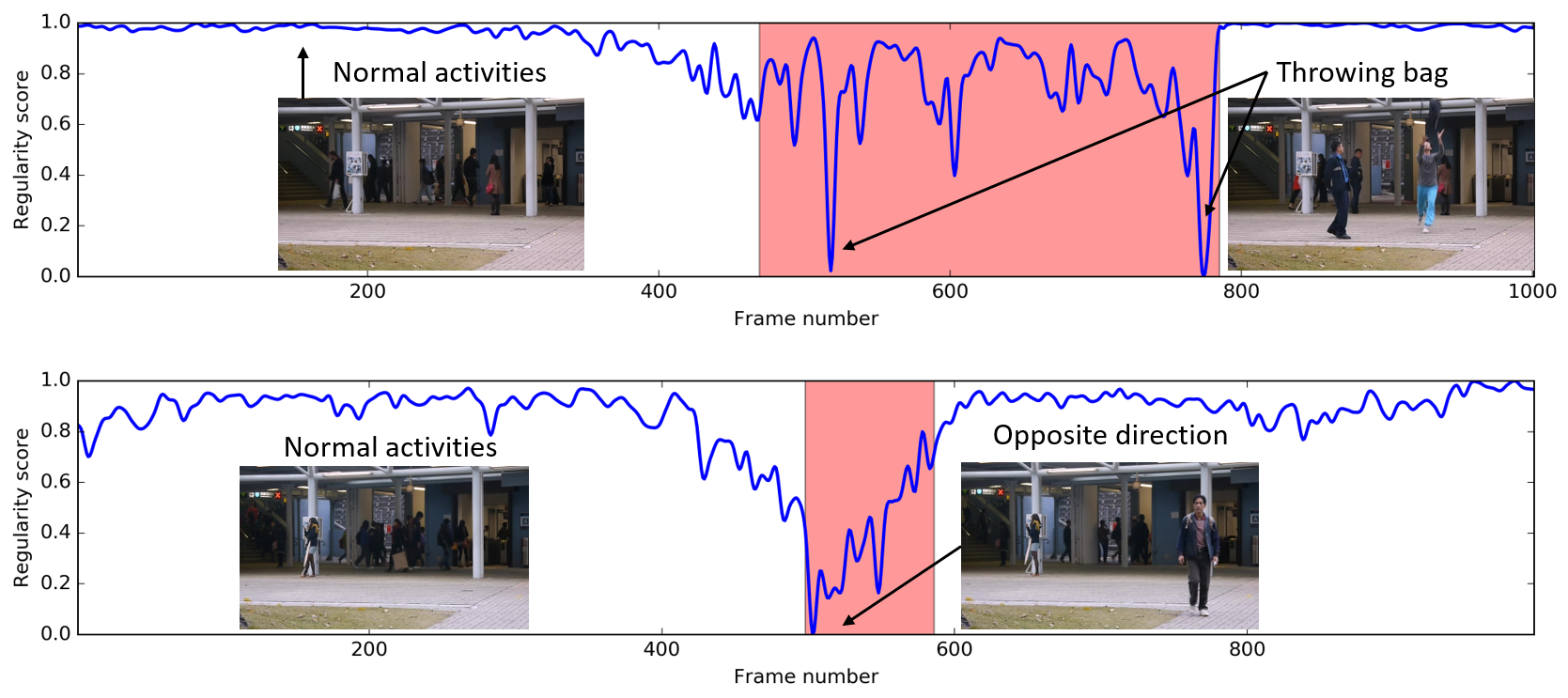}
	\caption{Regularity score of video \#5 (top) and \#15 (bottom) from the Avenue dataset.}
	\label{fig:avenue_show}
\end{figure}

\begin{figure}
	\centering
	\includegraphics[width=\textwidth]{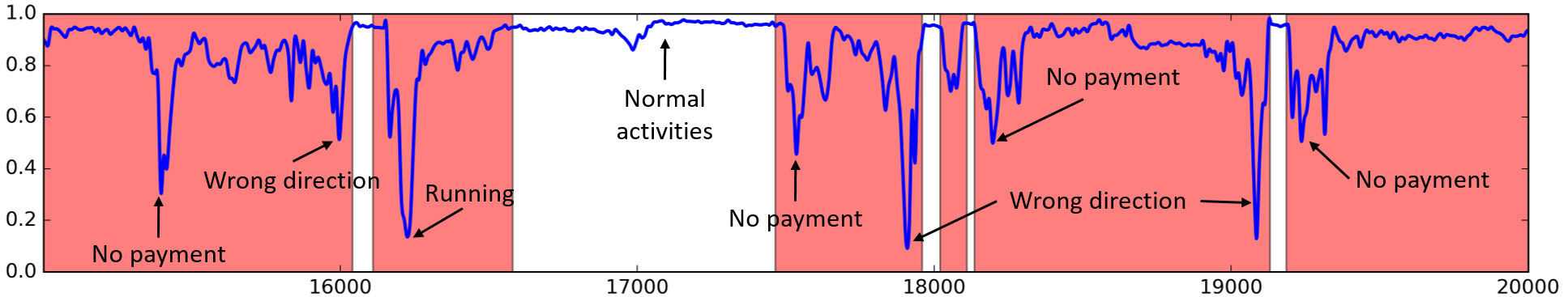}
	\caption{Regularity score of frames 115000-120000 from the Subway Entrance video.}
	\label{fig:enter_show}
\end{figure}

\begin{figure}
	\centering
	\includegraphics[width=\textwidth]{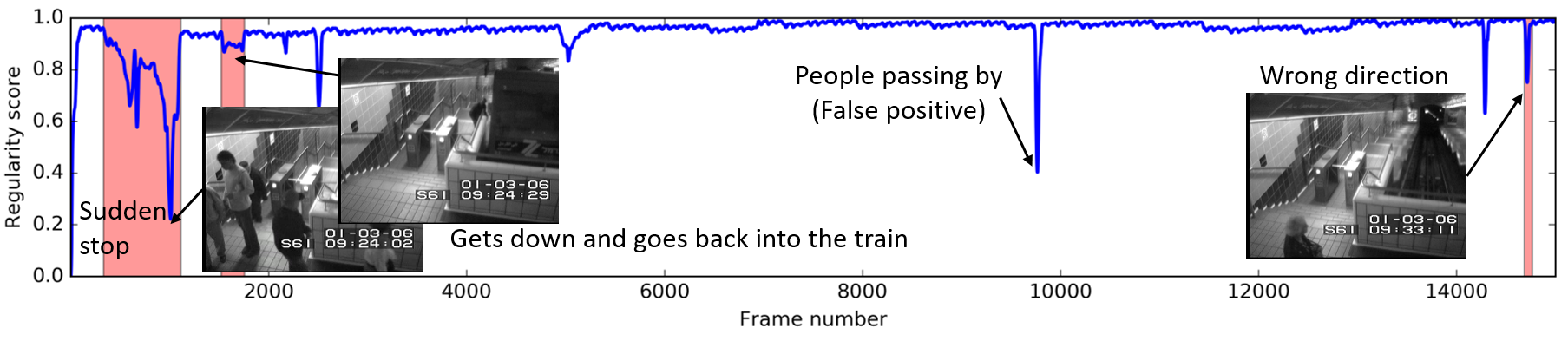}
	\caption{Regularity score of frames 22500-37500 from the Subway Entrance video.}
	\label{fig:exit_show}
\end{figure}

\begin{figure}
	\centering
	\includegraphics[width=\textwidth]{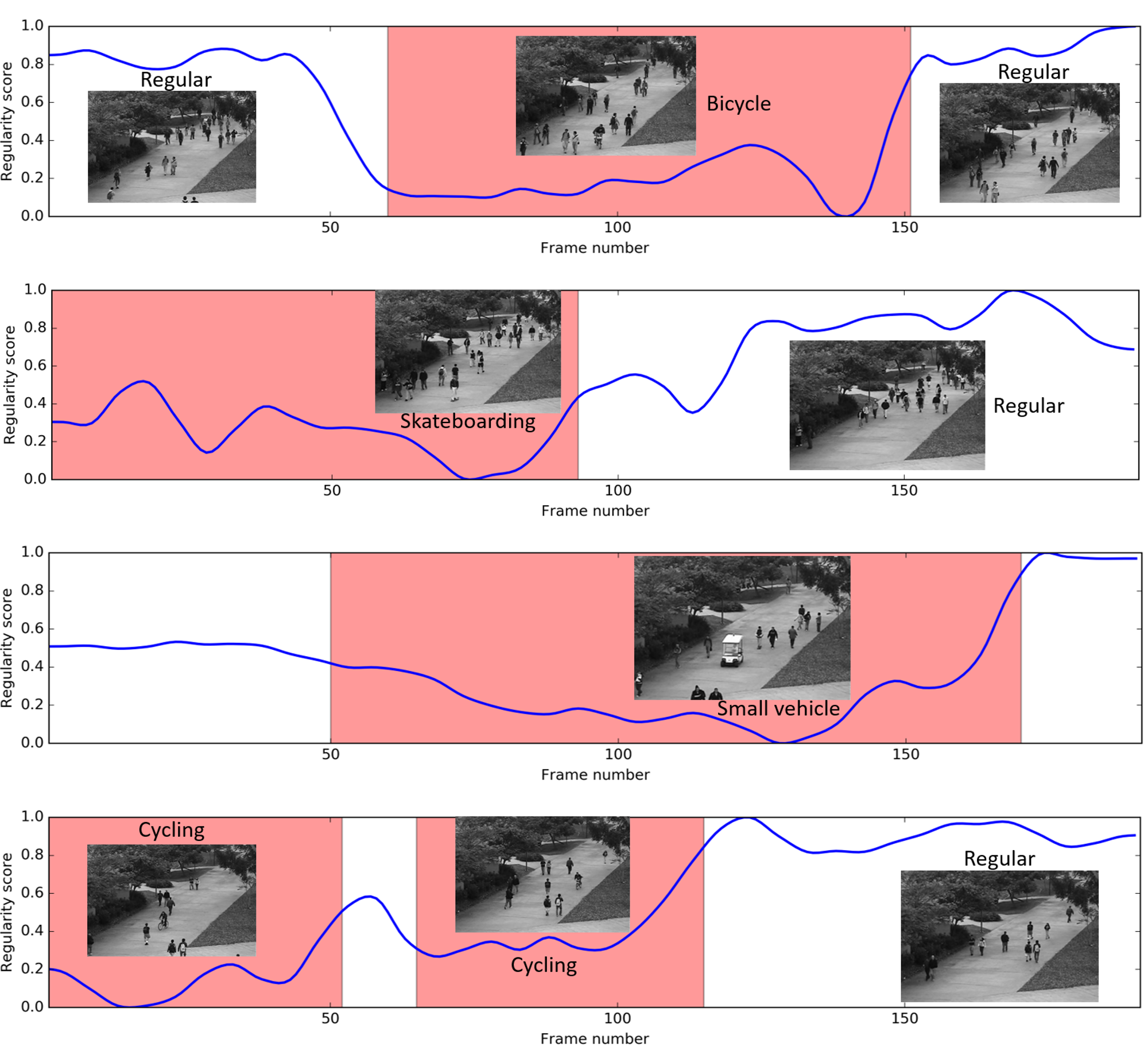}
	\caption{Regularity score of video \#1, \#8, \#24 and \#32 (from top to bottom) from UCSD Ped1 dataset.}
	\label{fig:ped1_show}
\end{figure}

\begin{figure}
	\centering
	\includegraphics[width=\textwidth]{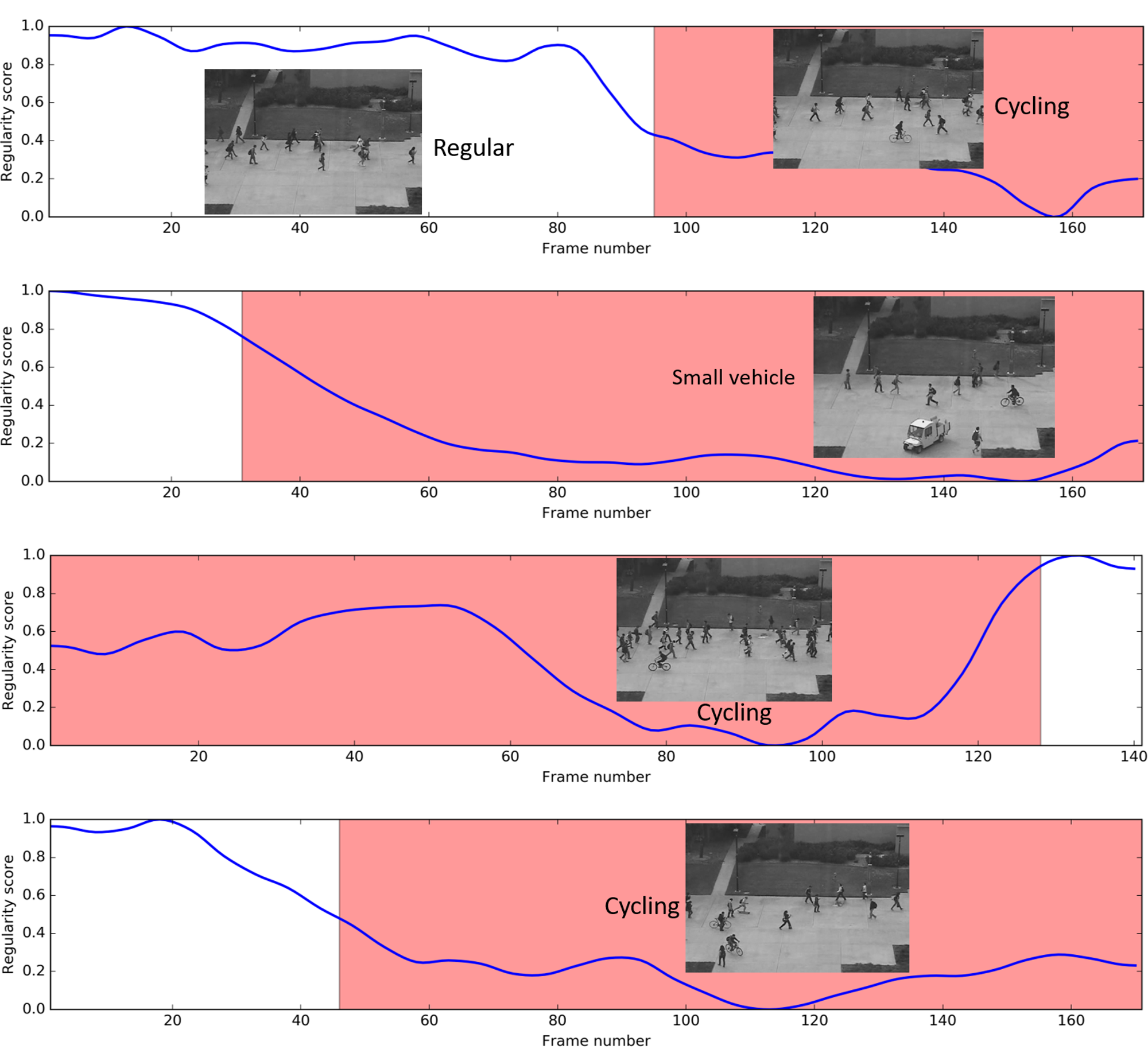}
	\caption{Regularity score of video \#2, \#4, \#5 and \#7 (from top to bottom) from UCSD Ped2 dataset.}
	\label{fig:ped2_show}
\end{figure}

Figure \ref{fig:avenue_show}, \ref{fig:enter_show}, and \ref{fig:exit_show} illustrate the output of the proposed system on samples of the Avenue dataset, Subway entrance and exit scenes respectively; our method detects anomalies correctly in these cases even in crowded scenes. 

Almost all anomalies produce strong downward spikes which indicate a low regularity score, including a difficult-to-detect skateboarding activity as illustrated in Figure \ref{fig:ped1_show}.

\subsubsection{Comparing Our Method with 2D Convolutional Autoencoder (ConvAE)}

\begin{figure}
	\centering
	\includegraphics[width=\textwidth]{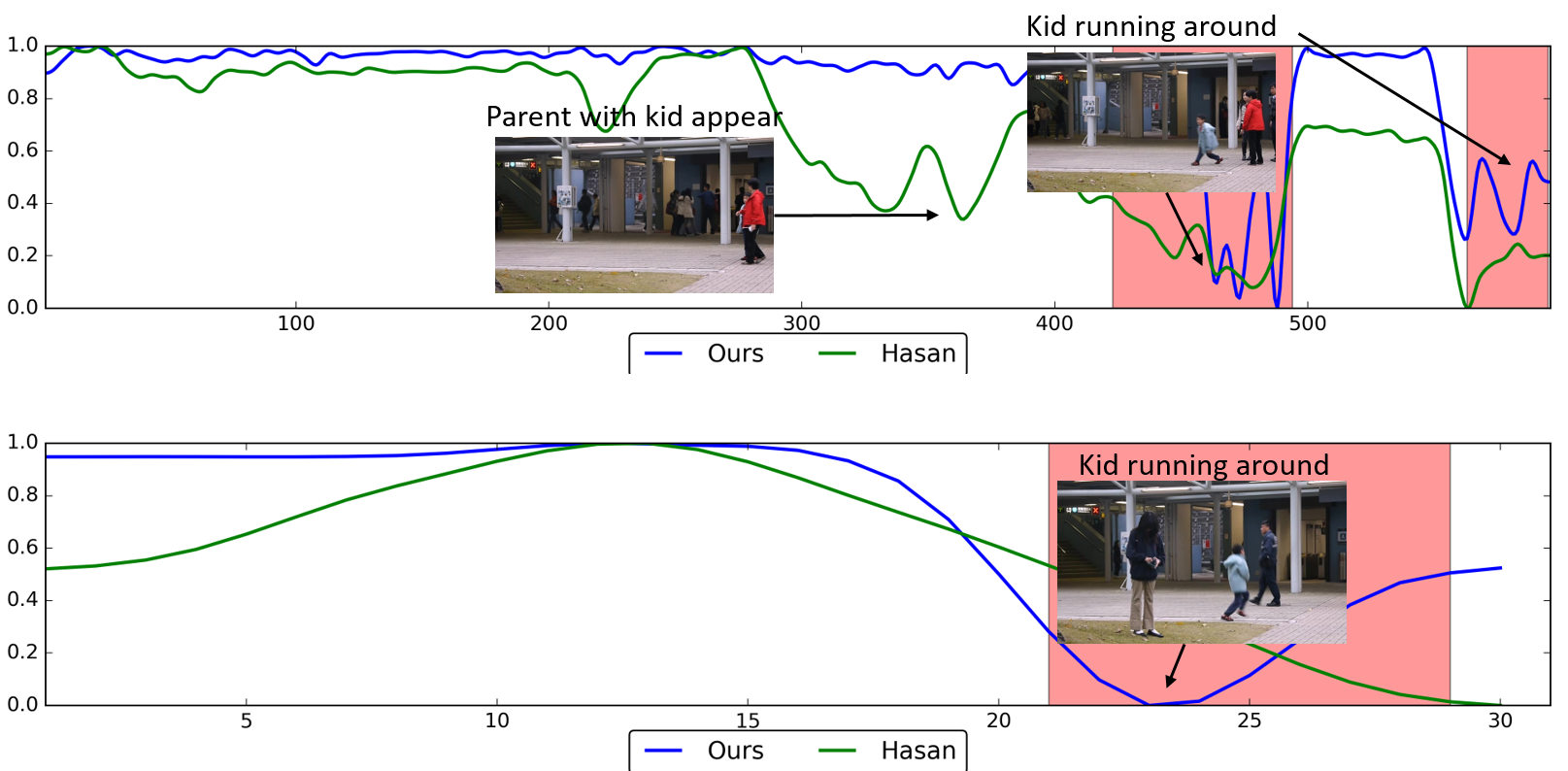}
	\caption{Comparing our method with ConvAE \cite{hasan2016} on Avenue dataset video \#7 (top) and \#8 (bottom). Best viewed in colour.}
	\label{fig:avenue_compare}
\end{figure}

\begin{figure}
	\centering
	\includegraphics[width=\textwidth]{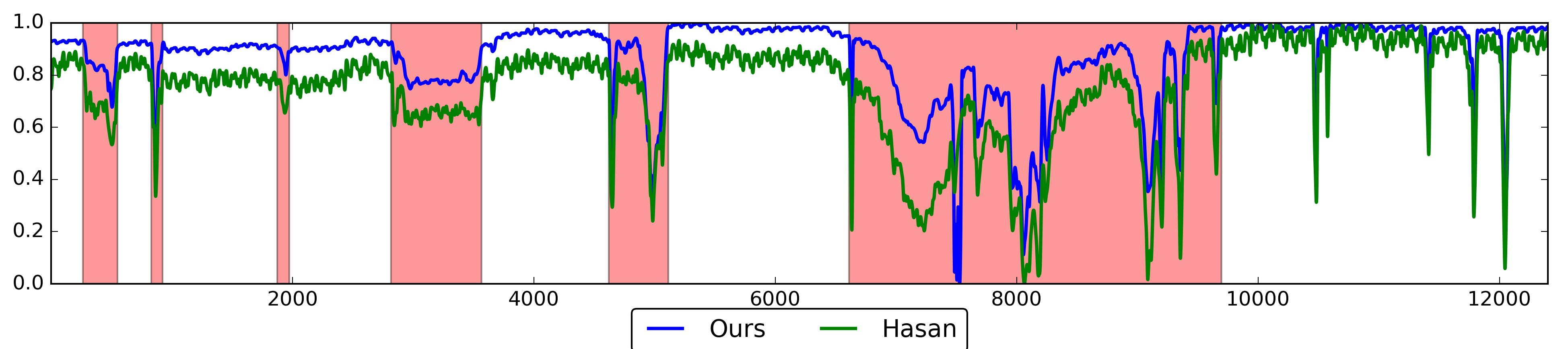}
	\caption{Comparing our method with ConvAE \cite{hasan2016} on Subway Exit video frames 10000-22500. Best viewed in colour.}
	\label{fig:exit_compare}
\end{figure}

\begin{figure}
	\centering
	\includegraphics[width=\textwidth]{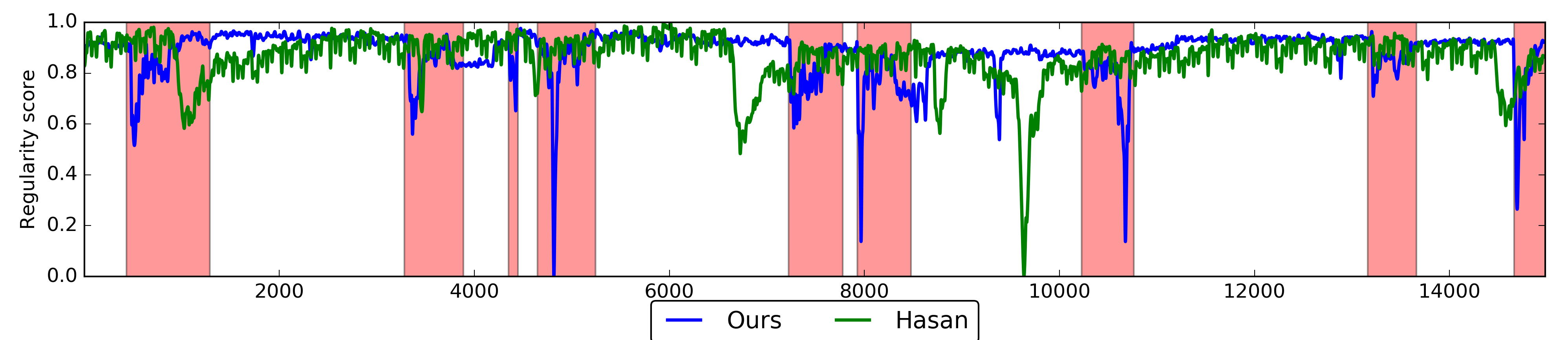}
	\caption{Comparing our method with ConvAE \cite{hasan2016} on Subway Entrance video frames 120000-144000. Best viewed in colour.}
	\label{fig:enter_compare}
\end{figure}

From Figure \ref{fig:avenue_compare} and \ref{fig:exit_compare}, it is easy to see that our method has detected more abnormal events with fewer false alarms compared to \cite{hasan2016}. As observed in Figure \ref{fig:enter_compare}, our method is able to produce higher regularity score during normal activities and lower scores when there are abnormalities.

\section{Conclusion}

In this research, we have successfully applied deep learning to the challenging video anomaly detection problem. We formulate anomaly detection as a spatiotemporal sequence outlier detection problem and applied a combination of spatial feature extractor and temporal sequencer ConvLSTM to tackle the problem. The ConvLSTM layer not only preserves the advantages of FC-LSTM but is also suitable for spatiotemporal data due to its inherent convolutional structure. By incorporating convolutional feature extractor in both spatial and temporal space into the encoding-decoding structure, we build an end-to-end trainable model for video anomaly detection. The advantage of our model is that it is semi-supervised -- the only ingredient required is a long video segment containing only normal events in a fixed view. Despite the model’s ability to detect abnormal events and its robustness to noise, depending on the activity complexity in the scene, it may produce more false alarms compared to other methods. For future work, we will investigate how to improve the result of video anomaly detection by active learning -- having human feedback to update the learned model for better detection and reduced false alarms. One idea is to add a supervised module to the current system, which the supervised module works only on the video segments filtered by our proposed method, then train a discriminative model to classify anomalies when enough video data has been acquired.

\bibliographystyle{splncs03} 
\bibliography{citations}

%\section*{Appendix: Demo Video}

%A sample demo video will be uploaded to YouTube in a few weeks.

\end{document}